\RequirePackage{fix-cm}
\documentclass[twocolumn]{svjour3}          
\smartqed  
\usepackage{graphicx}
\usepackage{multirow}
\usepackage[switch]{lineno}
\usepackage{color}
\usepackage[utf8]{inputenc}
\usepackage{blindtext, graphicx,nicefrac}
\usepackage{algorithm}
\usepackage{algpseudocode}
\usepackage{xurl}
\usepackage{soul}

\usepackage{amsthm,amsmath,amssymb,mathrsfs}
\usepackage{relsize}
\usepackage[table,xcdraw]{xcolor}
\usepackage{subfig}
\usepackage{flushend}
\usepackage{csquotes}
\usepackage{booktabs}
\usepackage{comment}
\usepackage{gensymb}
\usepackage{enumerate}
\usepackage{enumitem}
\usepackage{tikz}
\usepackage{footmisc}
\usepackage{lineno}
\usepackage{etoolbox}
\newcommand*{\affmark}[1][*]{\textsuperscript{#1}}
\usepackage[misc,geometry]{ifsym} 

\usepackage{tikz,xcolor,hyperref}

\definecolor{lime}{HTML}{A6CE39}
\DeclareRobustCommand{\orcidicon}{%
	\begin{tikzpicture}
	\draw[lime, fill=lime] (0,0) 
	circle [radius=0.16] 
	node[white] {{\fontfamily{qag}\selectfont \tiny ID}};
	\draw[white, fill=white] (-0.0625,0.095) 
	circle [radius=0.007];
	\end{tikzpicture}
	\hspace{-2mm}
}

\foreach \x in {A, ..., Z}{%
	\expandafter\xdef\csname orcid\x\endcsname{\noexpand\href{https://orcid.org/\csname orcidauthor\x\endcsname}{\noexpand\orcidicon}}
}

\journalname{Journal}
\begin{document}
\sloppy

\title{Differentially Private Multivariate Time Series Forecasting of Aggregated Human Mobility With Deep Learning: Input or Gradient Perturbation?}

\titlerunning{Differentially Private Time Series Forecasting of Human Mobility with DL}

\authorrunning{H. H. Arcolezi, J.F. Couchot, D. Renaud \textit{et al.}} 

\author{Héber H. Arcolezi\affmark[1]\textsuperscript{,}\affmark[2]\orcidA{}          \and
        Jean-François Couchot\affmark[2]\orcidB{}    \and
        Denis Renaud\affmark[3]\orcidC{}            \and
        Bechara Al Bouna\affmark[4]\orcidD{}        \and
        Xiaokui Xiao\affmark[5]\orcidE{}
}

\institute{ \Letter  \text{ } Héber H. Arcolezi \\
        \hspace*{0.4cm}  heber.hwang-arcolezi@inria.fr \\ 
        \\
\affmark[1]Inria and École Polytechnique (IPP), Palaiseau, France\\
\affmark[2]Femto-ST Institute, Univ. Bourg. Franche-Comt\'e, UBFC, CNRS, Belfort, France\\
\affmark[3]Orange Applications For Business, Orange Labs., Belfort, France \\
\affmark[4]TICKET Lab., Antonine University Hadat-Baabda, Baabda, Lebanon\\
\affmark[5]School of Computing, National University of Singapore, Singapore, Singapore
}

\date{Final version accepted in the journal Neural Computing and Applications. \\Version of Record: \url{https://doi.org/10.1007/s00521-022-07393-0}}

\maketitle
\begin{abstract}
This paper investigates the problem of forecasting multivariate aggregated human mobility while preserving the privacy of the individuals concerned. Differential privacy, a state-of-the-art formal notion, has been used as the privacy guarantee in two different and independent steps when training deep learning models. On one hand, we considered \textit{gradient perturbation}, which uses the differentially private stochastic gradient descent algorithm to guarantee the privacy of each time series sample in the learning stage. On the other hand, we considered \textit{input perturbation}, which adds differential privacy guarantees in each sample of the series before applying any learning. We compared four state-of-the-art recurrent neural networks: Long Short-Term Memory, Gated Recurrent Unit, and their Bidirectional architectures, i.e., Bidirectional-LSTM and Bidirectional-GRU. Extensive experiments were conducted with a real-world multivariate mobility dataset, which we published openly along with this paper. As shown in the results, differentially private deep learning models trained under gradient or input perturbation achieve nearly the same performance as non-private deep learning models, with loss in performance varying between $0.57\%$ to $2.8\%$. The contribution of this paper is significant for those involved in urban planning and decision-making, providing a solution to the human mobility multivariate forecast problem through differentially private deep learning models.

\keywords{Mobility prediction \and Differential privacy \and Crowd flow \and Differentially private machine learning.}
\end{abstract}

\section{Introduction}\label{sec:intro}

Efficiently planning a road network, choosing the optimal location for a hospital, for example, are all decisions based on a precise understanding of human mobility. Mobile phone data such as call detail records (CDRs) have proven to be one of the most promising ways to analyze human mobility on a large scale due to the high penetration rates of cell phones~\cite{deMontjoye2018,Buckee2020,Blondel2015}. CDR is a type of metadata that describes users' activities in a cellular network (e.g., phone calls, SMS) with information such as the duration of communication, the antennas that handled the service (coarse level location), and so on. For this reason, CDRs are commonly used by mobile network operators (MNOs) to enhance their services and for billing and legal purposes~\cite{Oliver2020,Blondel2015}. 

Because both temporal and spatial information is available in CDRs, these data have become one of the most important data sources for research on human mobility~\cite{Buckee2020,luca2020deep,deMontjoye2018,deAlarcon2021,Dujardin2020}. Indeed, human mobility analysis can benefit individuals and society enabling local authorities to improve urban planning, enhance the transportation system, and assist in decision-making to respond to critical situations (e.g., natural disasters~\cite{Dujardin2020,Hong2018}). Within a recent context, due to the ongoing Coronavirus Disease 2019 (COVID-19) pandemic~\cite{who_annouces_pandemic}, on 8 April 2020, the European Commission asked MNOs in the European region to share anonymized and aggregated mobility data to help to fight the outbreak~\cite{Vespe2021,europecom2020}, which has also been done in other parts of the world as described in~\cite{deAlarcon2021}. This vision is also shared by, e.g., Buckee et al.~\cite{Buckee2020} and Oliver et al.~\cite{Oliver2020}, which highlight the importance of aggregate mobility data and mobile phone data like CDRs for fighting the COVID-19 outbreak. 

\begin{figure*}[t]
    \centering
    \includegraphics[width=0.835\linewidth]{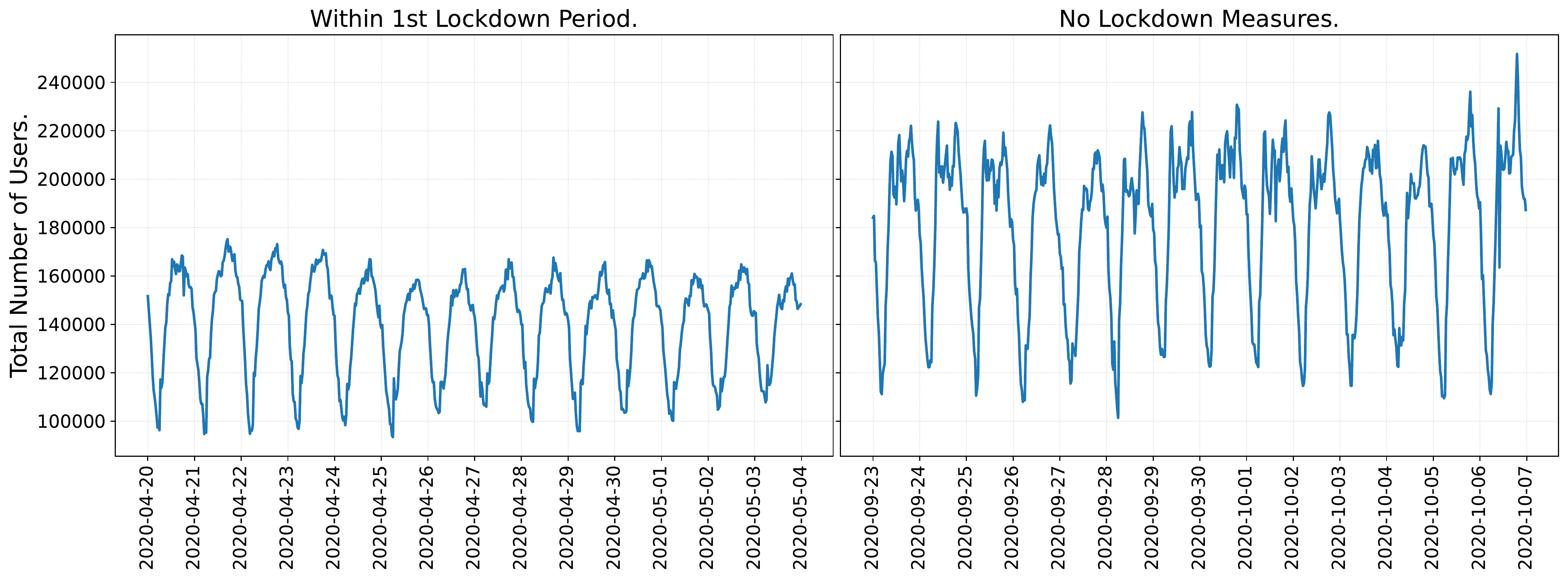}
    \caption{Aggregated human mobility analytics in Paris during the COVID-19 pandemic: two weeks within the first lockdown period in France (left-side plot) and during two weeks with no lockdown measures (right-side plot).}
    \label{fig:analysis_lockdown}
\end{figure*}

For instance, on analyzing the dataset at our disposal (further explained in Subsection~\ref{sub:dataset_problem_statement}), Fig.~\ref{fig:analysis_lockdown} illustrates aggregated mobility analytics in Paris, France, for two 14-days periods: from the beginning of 2020-04-21 to the end of 2020-05-03 and from the beginning of 2020-09-23 to the end of 2020-10-06. The plot on the left-side corresponds to mobility analytics during the first national lockdown period in France~\cite{lockdown}, and the plot on the right-side corresponds to a period with no lockdown measures. As one can notice, there is a clear difference between the first period of analysis (low mobility activity) and the second one (high mobility activity). This type of mobility analysis provides important insights on mobility patterns for public authorities and policymakers, for example~\cite{Vespe2021,deAlarcon2021}. 

However, on analyzing mobility data, some studies have shown that humans follow particular patterns with a high identifiability~\cite{deMontjoye2013,Murakami2021} and, hence, \textit{users' location privacy is a major concern}~\cite{deMontjoye2013,Murakami2021,Arcolezi2021,deMontjoye2018,app_blip,Acs2014}. Indeed, even though in CDRs the location information is at a coarse level (antennas that handled the service), collecting many imprecise locations can still lead to privacy leaks, such as the home or work addresses. Also, this is a scenario in which users cannot sanitize their data locally since CDRs are automatically generated on MNOs' servers through the use of a service (e.g., making/receiving phone calls). To tackle these issues, the General Data Protection Regulation (GDPR)~\cite{GDPR} as well as some data protection authorities, such as the Commission Nationale de l'Informatique et des Libertés (CNIL)~\cite{CNIL}, in France, require that MNOs anonymize ``on-the-fly" CDRs used for purposes other than billing. More precisely, if CDRs are used for mobility analytics, these data must be processed within a required time interval (e.g., 15 minutes) if and only if there is a sufficient number of users present for reaching a specific level of anonymity (i.e., ``hide in the crowd").

This way, MNOs tend to publish aggregated mobility data~\cite{deAlarcon2021,Xu2017,Vespe2021,Tu2018,fluxvision1}, e.g., the number of users by coarse location at a given timestamp, which, in other words, represents a \textit{multivariate time series dataset} that can be used for predictive mobility~\cite{luca2020deep}. Nevertheless, as recent studies have shown, even aggregated mobility data can be subject to membership inference attacks~\cite{Pyrgelis2017,Pyrgelis2020} and users' trajectory recovery attack~\cite{Tu2018,Xu2017}, thus requiring proper sanitization. To tackle privacy concerns in data releases, research communities have proposed different methods to preserve privacy, with differential privacy (DP)~\cite{Dwork2006,dwork2014algorithmic} standing out as a formal definition that allows quantifying the privacy-utility trade-off. Differential privacy has also been at the core of many privacy-preserving machine learning (ML) and deep learning (DL) models~\cite{Shokri2015,DL_DP,pytorch_privacy,chaudhuri2011differentially,Chamikara2020,tf_privacy} since predictive models are also subject to privacy attacks~\cite{Carlini2021,yang2021privacy,Song2017,Carlini2019,shokri2017membership}.

With these elements in mind, this paper contributes with a comparative analysis between adding DP guarantees into two different steps of training DL models to forecasting multivariate aggregated human mobility data. On the one hand, we consider using \textit{gradient perturbation}, which can be achieved by training DL models over original time-series data with the differentially private version of the stochastic gradient descent algorithm (DP-SGD)~\cite{DL_DP}. On the other hand, we consider using \textit{input data perturbation}, i.e., training DL models with differentially private time series data. Notice that, while aggregated time-series data provides some anonymity-based protection, with the latter input perturbation setting, DP also adds a layer of protection against, e.g., data breaches~\cite{data_breaches}, membership inference attacks~\cite{Pyrgelis2017,Pyrgelis2020}, and users' trajectory recovery attacks~\cite{Tu2018,Xu2017}.

It is worth mentioning that human mobility forecast information is of great importance for public and/or private organizations to identify strategies to propose better decision-making solutions for society~\cite{ebola,deMontjoye2018,deAlarcon2021,Oliver2020,Dujardin2020,Hong2018,Buckee2020,Vespe2021}. Therefore, in this paper, extensive experiments were carried out with a real-world mobility dataset collected by Orange~\cite{fluxvision1} on analyzing CDRs in 6 coarse regions in Paris, France, which we publish openly as an open mobility dataset. More precisely, this paper benchmarks four state-of-the-art DL models with this dataset, i.e., recurrent neural networks (RNNs): Long Short-Term Memory (LSTM)~\cite{LSTM}, which is capable of learning long-term dependencies while overcoming the vanishing gradient problem of standard RNNs; Gated Recurrent Unit (GRU)~\cite{GRU}, which is similar to LSTM but with a simpler architecture; and their Bidirectional~\cite{BI_RNN} architectures, i.e., BiLSTM and BiGRU. Moreover, we also took into consideration users' privacy, adding DP guarantees into the predictive models and evaluating their utility loss in comparison with non-private DL models.

\textbf{To summarize, this paper makes the following contributions:}

\begin{itemize}
    \item Publish the real-world, CDRs-based, and multivariate (i.e., 6 coarse regions) aggregated mobility dataset openly in a Github repository\footnote{\url{https://github.com/hharcolezi/ldp-protocols-mobility-cdrs}}.
    \item Benchmark four state-of-the-art RNNs (LSTM, GRU, BiLSTM, and BiGRU) with this dataset for \textit{one-step-ahead} multivariate forecasting.
    \item Provide the first comparative evaluation on the impact of differential privacy guarantees when training DL models in both input and gradient perturbation settings, for multivariate time series forecasting.
\end{itemize}

Therefore, we intend that from this study, other classical time series forecasting, ML, and privacy-preserving ML techniques can be tested and compared. 

\noindent \textbf{Outline.} The remainder of this paper is organized as follows. In Section~\ref{sec:data_metho}, we describe the material and methods used in this work, i.e., the mobility data and the problem statement; the DL methods, and the privacy guarantee, namely, differential privacy for both input and gradient perturbation settings. In Section~\ref{sec:results}, we present the experimental setup, our results with non-private DL models, and our results with differentially private DL models. In Section~\ref{sec:discussion}, we discuss our work with its limitations and future directions. Lastly, in Section~\ref{sec:conclusion}, we present the concluding remarks.

\section{Material and Methods} \label{sec:data_metho}

In this section, we first describe the mobility dataset and the problem we intend to solve (Subsection~\ref{sub:dataset_problem_statement}). Next, we briefly describe the DL methods we consider in our experiments (Subsection~\ref{sub:dl}). Lastly, we recall the privacy notion that we are considering, i.e., differential privacy, in both input and gradient perturbation settings (Subsection~\ref{sub:dp}).

\subsection{Mobility Dataset and Problem Statement} \label{sub:dataset_problem_statement}

The dataset at our disposal was provided by an MNO in France~\cite{fluxvision1}, which contains anonymized and aggregated human mobility data resulted from analyzing CDRs ``on-the-fly", following recommendations from both GDPR and CNIL. This dataset comprises information for two periods: from 2020-04-20 to 2020-05-03 and from 2020-08-24 to 2020-11-04, with time granularity of $30$ minutes (min) and spatial granularity of 6 coarse regions in Paris, France. 

More formally, this is a multivariate time series dataset $X_{(t_1,t_{\tau})}$ with aggregate number of people per region and corresponding time period $t \in [1, \tau]$. That is, $X_{(t_1,t_{\tau})} = [\langle t_1, \textbf{x}_{1}\rangle, \langle t_2, \textbf{x}_{2} \rangle, ..., \langle t_{\tau}, \textbf{x}_{\tau} \rangle ]$, in which $\textbf{x}_{t}$ is a vector with each position representing the number of users per region at time $t \in[1,\tau]$. In this paper, we aim at forecasting the future number of people at the next $30$-min interval in each of the 6 regions. Thus, given $X_{(t_1,t_{\tau})}$, the goal is to forecast $X_{(t_{\tau + 1})}$, i.e., \textit{one-step-ahead} forecasting, which is unknown at time $\tau$.

For the rest of this paper, we only utilize the second period of this dataset (i.e., from 2020-08-24 to 2020-11-04), which has aggregated mobility data for $72$ days. For each week, coarse region, and $30$-min interval, we used the interquartile range technique\footnote{\url{https://en.wikipedia.org/wiki/Interquartile\_range}} to detect outliers and missing data. These values were completed with the average value for that respective week, coarse region, and $30$-min interval. Table~\ref{tab:statistics_data} presents descriptive statistics about this processed dataset with the following measures per region (labeled as R1 - R6): min, max, mean, standard deviation (std), and median. 

\begin{table*}
    \centering
    \caption{Descriptive statistics for the multivariate time series dataset on the number of users per coarse region.}
    \begin{tabular}{c c c c c c c}
    \hline
      \textbf{Statistic} &\textbf{R1}  &\textbf{R2}   &\textbf{R3}   &\textbf{R4}   &\textbf{R5}   &\textbf{R6}   \\\hline
        Min &   56,937 &   1,996 &   1,429 &    255 &   252 &    347 \\\hline
        Max &  165,405 &  21,980 &  28,990 &  25,184 &  7,961 &  27,637 \\\hline
        Mean &  116,777 &  14,307 &  16,274 &  11,758 &  4,166 &  11,559 \\\hline
        Std &   17,947 &   2,803 &   3,915 &   3,682 &  1,450 &   5,136 \\\hline
        Median &  121,488 &  14,808 &  16,661 &  12,134 &  4,495 &  12,542 \\\hline
    \end{tabular}
    \label{tab:statistics_data}
\end{table*}

\subsection{Deep Learning Methods} \label{sub:dl}

To predict the number of users in each coarse region in a multivariate time series forecasting framework, we compared the performance of four state-of-the-art RNNs: LSTM~\cite{LSTM}, GRU~\cite{GRU}, and their Bidirectional~\cite{BI_RNN} architectures, i.e., BiLSTM and BiGRU. Indeed, RNNs is a specialized class of neural networks used to process sequential data (e.g., time-series data). RNNs have at least one feedback connection that provides the ability to use contextual information when mapping between input and output sequences~\cite{Hewamalage2021}. The LSTM, GRU, and Bidirectional RNN methods are briefly described in the following subsections.

\subsubsection{Long Short-Term Memory} \label{subsub:lstm}
Long Short-Term Memory~\cite{LSTM} is a type of RNN that overcomes the vanishing gradient problem of standard RNNs. The memory cell of LSTM divides its states in long-term state \(c_{(t)}\) and short-term state \(h_{(t)}\). The learning process is controlled by three gates: input \(i_{(t)}\), forget \(f_{(t)}\) and output \(o_{(t)}\) gates, which give LSTM the ability to learn which data in a sequence is important to keep or to discard. More precisely, both input \(x_{(t)}\) and the previous short-term state \(h_{(t-1)}\) are fed to four different and fully connected layers. Then, the first layer computes the internal hidden state \(g_{(t)}\), using \(x_{(t)}\) and \(h_{(t-1)}\), and partially store \(g_{(t)}\) in the long-term state. The remaining tree layers are the nonlinear gating units. The forget gate $f_{(t)}$ controls the past information which must be vanished or kept. The input gate $i_{(t)}$ controls the new information which is to be added to the long-term state. Lastly, the output gate $o_{(t)}$ controls which information could be utilized for the output of the memory cell $y_{(t)}$. The mathematical formulation is as follows~\cite{LSTM}:

\begin{equation*}
\begin{aligned}
i_{(t)} &= \sigma(W_{xi} x_{(t)} + W_{hi} h_{(t-1)} + b_i)\textrm{,}\\
f_{(t)} &= \sigma(W_{xf} x_{(t)} + W_{hf} h_{(t-1)} + b_f)\textrm{,}\\
o_{(t)} &= \sigma(W_{xo} x_{(t)} + W_{ho} h_{(t-1)} + b_o)\textrm{,}\\
g_{(t)} &= \tanh(W_{xg} x_{(t)} + W_{hg} h_{(t-1)} + b_g)\textrm{,}\\
c_{(t)} &= f_{(t)} \otimes c_{(t-1)} + i_{(t)} \otimes g_{(t)}\textrm{,}\\
y_{(t)} &= h_{(t)} = o_{(t)} \otimes \tanh(c_{(t)})\textrm{,}    
\end{aligned}
\end{equation*}

\noindent in which \(\otimes\) means an element-wise multiplication, $\sigma$ is the sigmoid function, $W_*$ are weight matrices, and $b_*$ are the vectors of bias term.

\subsubsection{Gated Recurrent Unit} \label{subsub:gru}
Gated Recurrent Unit~\cite{GRU} is also a type of RNN, which works using the same principle as LSTM. GRU utilizes two gates: update $z_{(t)}$ and reset $r_{(t)}$, which decide what information should be passed to the output. More specifically, the reset gate $r_{(t)}$ controls how to combine the new input with the previous memory. The update gate $z_{(t)}$ controls how much of the last memory to keep. The mathematical formulation is as follows~\cite{GRU}:

\begin{equation*}
\begin{aligned}
z_{(t)} &= \sigma(W_{z} x_{(t)} + U_{z} h_{(t-1)})      \textrm{,}\\
r_{(t)} &= \sigma(W_{r} x_{(t)} + U_{r} h_{(t-1)})      \textrm{,}\\
c_{(t)} &= \tanh(W x_{(t)} + U(r_{(t)} \otimes h_{(t-1)}))  \textrm{,}\\
h_{(t)} &= (1 - z_{(t)}) h_{(t-1)} + (z_{(t)} c_{(t)}) \textrm{.}    
\end{aligned}
\end{equation*}

\subsubsection{Bidirectional RNN} \label{subsub:bidirectional}    
Bidirectional RNN (BiRNN)~\cite{BI_RNN} is a combination of two RNNs: one RNN moves forward while the other moves backward. That is, BiRNN connects two hidden layers of opposite directions to the same output. The RNN cells in a BiRNN can either be standard RNNs, LSTMs, GRUs, and so on. This Bidirectional architecture allows the networks to have both backward and forward information about the sequence at every time step.

\subsection{Differential Privacy} \label{sub:dp}

In recent years, differential privacy~\cite{Dwork2006} has been increasingly accepted as the current standard for data privacy with several large-scale implementations in the real-world (e.g.~\cite{linkedin,aktay2020google}). One key reason is that DP addresses the paradox of learning about a population while learning nothing about single individuals~\cite{dwork2014algorithmic}. More specifically, the idea is that removing (or adding) a single row from the database should not affect \textit{much} the statistical results. A formal definition of DP is given in the following~\cite{dwork2014algorithmic}:

\begin{definition}[($\epsilon, \delta$)-Differential Privacy~\cite{dwork2014algorithmic}] Given $\epsilon>0$ and $0 \leq \delta <1$, a randomized algorithm ${\mathcal{A}: \mathcal{D} \rightarrow R}$ is said to provide {($\epsilon, \delta$)-differential-privacy (($\epsilon, \delta$)-DP}) if, for all neighbouring datasets $D_{1},D_{2} \in \mathcal{D}$ that differ on the data of one user, and for all sets $R$ of outputs:
\begin{equation}
{ \Pr[{\mathcal{A}}(D_{1})\in R]\leq e^{\epsilon } \Pr[{\mathcal{A}}(D_{2})\in R]} + \delta \textrm{.}
\label{eq:dp}
\end{equation}
\end{definition}

The additive $\delta$ is interpreted as a probability of failure. Normally, a common choice for $\delta$ is to set it significantly smaller than $1/n$ where $n$ is the number of users in the database~\cite{dwork2014algorithmic}. Throughout this paper, if $\delta=0$, we will just say that $\mathcal{A}$ is $\epsilon$-DP.

\subsubsection{Properties of Differential Privacy} \label{subsub:prop_dp}

Differential privacy possesses several important properties, highlighting its strength in comparison with other privacy models. For instance, with DP, there is no need to define the \textit{background knowledge} that attackers might have, which is equivalent to assuming an attacker with \textit{unlimited resources}. In addition, DP is immune to \textit{post-processing}, which means it is not possible to make an $\epsilon$-DP mechanism less differentially private by evaluating any function $f$ of the response of the mechanism, given that there is no additional information about the database.

\begin{proposition}[Post-Processing of DP~\cite{dwork2014algorithmic}] \label{prop:post_processing} If $\mathcal{A} : \mathcal{D} \rightarrow R$ is $\epsilon$-DP, then $f (\mathcal{A})$ is also $\epsilon$-DP for any function $f$.
\end{proposition}

Furthermore, DP also \textit{composes} well, which is one of the most powerful features of this privacy model. For instance, accounting for the \textit{overall} privacy loss consumed in a pipeline of several DP algorithms applied to the same database is feasible due to composition. We recall the sequential composition that will be used in this paper in the following.

\begin{proposition}[Sequential Composition~\cite{dwork2014algorithmic}] \label{prop:sequential_composition} Let $\mathcal{A}_1$ be an $\epsilon_1$-DP mechanism and $\mathcal{A}_2$ be an $\epsilon_2$-DP mechanism. Then, the mechanism $\mathcal{A}_{1,2}(\mathcal{D})=\left( \mathcal{A}_1(\mathcal{D}), \mathcal{A}_2(\mathcal{D})\right)$ is $(\epsilon_1+\epsilon_2)$-DP.
\end{proposition}

\subsubsection{Differentially Private Gaussian Mechanism}  \label{subsub:gauss_mech}

Any mechanism that respects Definition~\ref{eq:dp} can be considered differentially private. Two widely used DP mechanisms for numeric queries (i.e., functions $f : \mathcal{D} \rightarrow \mathbb{R}$) are the Laplace mechanism~\cite{Dwork2006} and the Gaussian mechanism~\cite{dwork2014algorithmic}. One important parameter that determines how accurately we can answer the queries is their \textit{sensitivity}. We recall the definition of $\ell_2$-sensitivity and the Gaussian mechanism that will be used in this paper in the following.

\begin{definition}[$\ell_2$-sensitivity~\cite{dwork2014algorithmic}] The $\ell_2$-sensitivity of a function $f : \mathcal{D} \rightarrow \mathbb{R}$, for all neighbouring datasets $D_{1},D_{2} \in \mathcal{D}$ that differ on the data of one user, is:
\begin{equation*}
    \Delta_2 (f) = max \textrm{  } || f(D_1) - f(D_2)  ||_2 \textrm{.}
\end{equation*}
\end{definition}

\begin{definition}[Gaussian mechanism~\cite{dwork2014algorithmic}] \label{def:gaussian_mech} For a query function $f : D \rightarrow \mathbb{R}$ over a dataset $D \in \mathcal{D}$ and for $\sigma = \frac{\Delta_2}{\epsilon} \sqrt{2 \ln{(1.25/\delta)}}$, the Gaussian mechanism is defined as:
\begin{equation*}
    \mathcal{A}_G (D, f(.), \epsilon, \delta) = f(D) + \mathcal{N}\left (0, \sigma^2 \right) \textrm{.}
\end{equation*}
\end{definition}

\noindent in which $\mathcal{N}\left (0, \sigma^2 \right)$ is the normal distribution centered at 0 with variance $\sigma^2$. For $\epsilon \in (0,1)$, the Gaussian mechanism provides ($\epsilon, \delta$)-DP~\cite{dwork2014algorithmic}.

\subsubsection{Differentially Private Deep Learning} \label{subsub:dp_dl}

In this paper, on the one hand, we consider that noise is added to each sample in the time series data (input perturbation). Once the data is differentially private, following Proposition~\ref{prop:post_processing}, any DL or pre-processing methods can be applied to the data. On the other hand, we consider the case where noise is added in the learning stage (gradient perturbation). In this case, the raw time-series data is used as input to a DL method trained with the differentially private stochastic gradient descent algorithm. These two settings are briefly described in the following:

\noindent \textbf{Input data perturbation.} Input perturbation (or data perturbation) consists to the fact that \textit{DP is added to each data sample} $\textbf{x}_i \in \mathcal{D}$. For example, let $\textbf{x}$ be a real-valued vector, then a differentially private version of it using the Gaussian mechanism (cf. Subsection~\ref{subsub:gauss_mech}) is: $\hat{\textbf{x}}=\textbf{x}+\mathcal{N}\left (0, \sigma^2 \right)$. On the one hand, input perturbation is the easiest method to apply and it is independent of any ML and post-processing techniques. On the other hand, the perturbation of each sample in the dataset may have a negative impact on the utility of the trained model. In this paper, we will use the Gaussian mechanism~\cite{dwork2014algorithmic} in Definition~\ref{def:gaussian_mech} to sanitize each release of multivariate mobility data.

\noindent \textbf{Gradient perturbation.} Another solution to guarantee DP to the trained model is perturbing intermediate values in iterative algorithms. In this case, the authors in~\cite{DL_DP} proposed a differentially private version of the stochastic gradient descent algorithm (i.e., DP-SGD). Indeed, DL models trained with DP-SGD~\cite{DL_DP} provide provable DP guarantees for their input data. Two new parameters are added to the standard stochastic gradient descent algorithm, namely, \textit{clip} and \textit{noise multiplier}. The former is used to bound how much each training point can impact the model's parameters and the latter is used to add controlled Gaussian noise to the clipped gradients in order to ensure DP guarantee to each data sample in the training dataset. In this paper, we use the DP-SGD implementation from the Tensorflow Privacy (TFP) library~\cite{tf_privacy} to implement our DL models.

\section{Experimental Results} \label{sec:results}

We divide this section in the following way. First, we describe general settings for our experiments (Subsection~\ref{sub:gen_setup}). Next, we present the development and evaluation of non-private DL models (Subsection~\ref{sub:non_private_DL}). Lastly, we present the development of differentially private DL models, which include both gradient and input perturbation settings (Subsection~\ref{sub:private_DL}).

\subsection{General Setup of Experiments} \label{sub:gen_setup}

\noindent \textbf{Environment.} All algorithms were implemented in Python 3.8.8 with Keras and TFP~\cite{tf_privacy} libraries.

\noindent \textbf{Temporal features.} We added the time of the day and the time of the week as cyclical features to help models recognizing low and high peak values of human mobility patterns.

\begin{figure*}[!ht]
    \centering
    \includegraphics[width=0.95\linewidth]{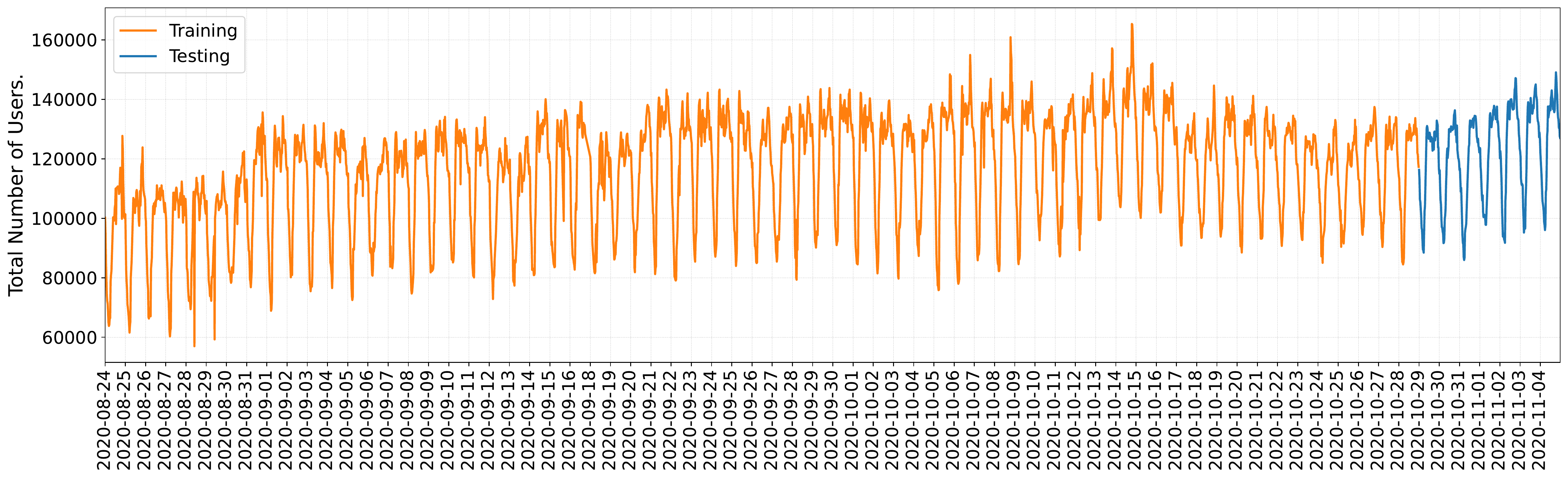}
    \caption{Example of data separation into training and testing sets for region R1.}
    \label{fig:train_test_split}
\end{figure*}

\noindent \textbf{Training and testing sets.} We split the dataset analyzed in Table~\ref{tab:statistics_data} from Subsection~\ref{sub:dataset_problem_statement} into exclusively divided learning (first $65$ days, i.e., $n_l=3120$ intervals of $30$-min) and testing (last $7$ days, i.e., $n_t=336$ intervals of $30$-min) sets. Fig.~\ref{fig:train_test_split} exemplifies the data separation into train and test sets for region R1.

\noindent \textbf{Forecasting methodology.} We used 6 prior time steps (i.e., lag values), which showed autocorrelation higher than 0.5 to predict a single step ahead in the future (i.e., short forecasting horizon). More specifically, the forecasting models will take into account the number of people in each coarse region from $3$ hours to make predictions \textit{one-step-ahead} for each coarse region in the next $30$-min interval. And in the end, we compute the performance metrics.

\noindent \textbf{Performance metrics.} All models were evaluated with standard time-series metrics, namely, root mean square error (RMSE) calculated as $RMSE=\frac{1}{n_t} \sqrt{\sum_{t=1}^{n_t} \left( y_t - \hat{y}_t \right)^2}$ and mean absolute error (MAE) calculated as $MAE=\frac{1}{n_t} \sum_{t=1}^{n_t} |y_t - \hat{y}_t|$; in which $y_t$ is the real output, $\hat{y}_t$ is the predicted output, and $n_t$ is the total number of samples in the \textit{test set}, for $t \in [1,n_t]$. RMSE was the primary metric to select the final DL models. As a multi-output scenario (i.e., $6$ coarse regions), we present the metrics per coarse region as well as its averaged values. In all experiments, due to randomness, we report the results of the best model over 10 runs. 

\subsection{Non-Private DL Forecasting Models} \label{sub:non_private_DL}

\noindent \textbf{Baseline model.} We compared the performance of the four DL methods (i.e., LSTM, GRU, BiLSTM, and BiGRU) with a naive forecasting technique a.k.a. ``\textit{persistence model}", which for each region, it returns the current number of people at time $t$ as the forecasted value, i.e., $\textbf{x}_{t+1}=\textbf{x}_t$. Notice that this is a quite accurate baseline since, in general, the number of people per coarse region varies slowly by 30-min (i.e., walking people may take more time to move from one coarse region to another).

\noindent \textbf{Model selection.} To select the best hyperparameters per DL method, we used Bayesian optimization~\cite{hyperopt2013} with $100$ iterations to minimize $loss=RMSE_{avg} + RMSE_{std}$; the subscripts \textit{avg} and \textit{std} indicates the averaged and standard deviation values of the RMSE metric considering the 6 coarse regions. For each method, we only used a single hidden layer followed by a dense layer (output), since RNNs generally perform well with a low number of hidden layers~\cite{Hewamalage2021}. So, we searched the following hyperparameters: number of neurons ($h_1$), batch size ($bs$), and learning rate ($\eta$). All models used ``relu" (rectified linear unit) as an activation function, which resulted in better performance than the default ``tanh" activation in prior tests. Lastly, models were trained using the adam (adaptive moment estimation) optimizer during $100$ epochs by minimizing the MAE loss function. Table~\ref{tab:hyper_non_private} exhibits the hyperparameters' search space and the final value used per DL method.

\begin{table*}[t]
    \centering
    \caption{Search space for hyperparameters and the best configuration obtained by each DL method.}
    \label{tab:hyper_non_private}
    \begin{tabular}{c c c c c c} 
    \hline
        \textbf{Hyperparameter's range} & \textbf{Step} & \textbf{LSTM} & \textbf{BiLSTM}  & \textbf{GRU}  & \textbf{BiGRU} \\ 
    \hline
         $h_1$: [25 -- 500]   & 25    &225     &500    &75    &175    \\
         $bs$: [5 -- 40]      & 5      &10     &10     &5    &5   \\
         $\eta$: [1e-5 -- 3e-3]  & --  &0.002233     &0.002303     &0.001725    &0.000289  \\
    \hline
    \end{tabular}
\end{table*}

\begin{figure*}[!ht]
    \centering
    \includegraphics[width=0.775\linewidth]{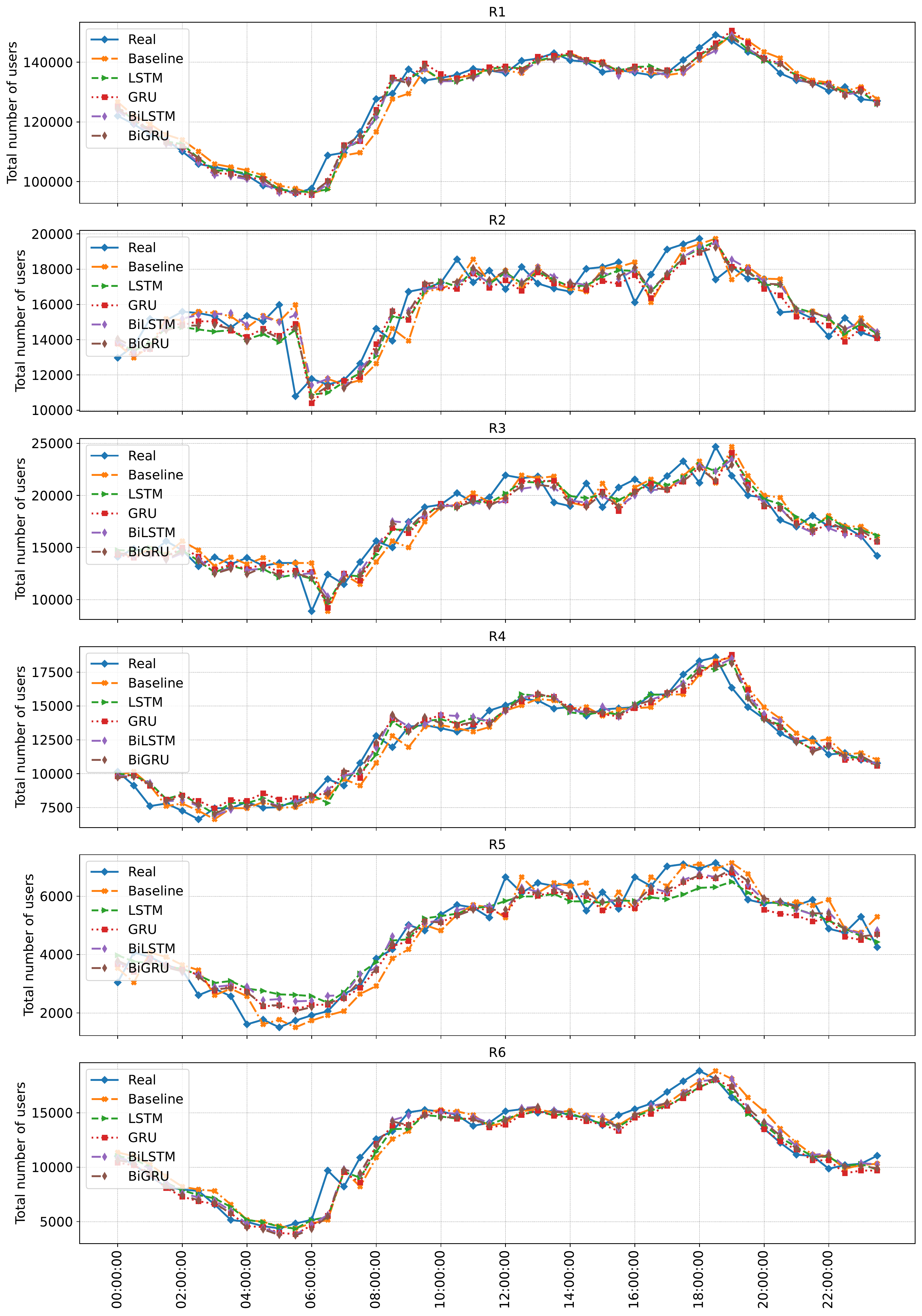}
    \caption{Multivariate time series forecast for the last day of the test set for the number of users per coarse region (R1 -- R6) by the following models: Baseline, LSTM, GRU, BiLSTM, and BiGRU.}
    \label{fig:results_pred_non_private}
\end{figure*}

\noindent \textbf{Results and analysis.} Table~\ref{tab:results_non_private} presents the performance of the developed DL models in comparison with the Baseline model based on RMSE and MAE metrics per region and the resulting mean. Notice that the metrics are in the real scale according to the number of users per region (cf. Table~\ref{tab:statistics_data}). That said, although R1 presents higher metric values, it does not necessarily mean worse results. One solution could be normalizing the data. Besides, Fig.~\ref{fig:results_pred_non_private} illustrates for each region forecasting results for the last day of our testing set, which includes the real number of people and the predicted ones by each RNN: LSTM, GRU, BiLSTM, and BiGRU.

\begin{table*}[t]
    \centering
    \caption{Performance of the Baseline model and non-private DL models based on RMSE and MAE metrics per region and the resulting mean values.}
    \label{tab:results_non_private}
    \begin{tabular}{c c c c c c c c c} 
    \hline
    \textbf{Model} & \textbf{Metric} & \textbf{R1} & \textbf{R2} & \textbf{R3}  & \textbf{R4}  & \textbf{R5}   & \textbf{R6}  & \textbf{Mean}  \\ 
    \hline
    \multirow{2}{*}{Baseline}   & RMSE &3461.6 &  1131.8 &  1517.9 &  986.5 &  561.3 &  1362.3  &1503.6    \\
                                & MAE  &2597.5 &   839.4 &  1105.8 &  744.1 &  434.3 &   921.5  &1107.1    \\ \hline
    \multirow{2}{*}{LSTM}       & RMSE &2667.2    &1007.3    &1291.6    &887.2    &536.3    &1135.6    &1254.2    \\
                                & MAE  &2053.8    &758.1    &969.8    &662.6    &432.3    &786.0    &943.8    \\ \hline
    \multirow{2}{*}{BiLSTM}     & RMSE &2572.7    &1033.3    &1276.4    &872.7    &528.1    &1166.7    &1241.6     \\
                                & MAE  &1954.7    &781.5    &965.5    &660.8    &419.4    &808.2    &931.7    \\ \hline
    \multirow{2}{*}{GRU}        & RMSE &2539.1    &973.0     &1296.0     &953.5    &499.9    &1185.1    &1241.1    \\
                                & MAE  &1949.7    &722.8    &939.6    &740.2    &396.4    &829.1    &929.6    \\ \hline
    \multirow{2}{*}{BiGRU}      & RMSE &2560.3    &968.3    &1282.6    &832.1    &478.9    &1163.7    &\textbf{1214.3}    \\
                                & MAE  &1957.2    &717.0    &955.3    &623.0    &382.7    &807.5    &\textbf{907.1}    \\ \hline 
    \end{tabular}
\end{table*}

As one can notice, all DL models consistently outperform the Baseline model. On average, the BiGRU model outperformed all other forecasting methods, with results highlighted \textbf{in bold} in Table~\ref{tab:results_non_private}. Indeed, for each region, the BiGRU consistently and considerably outperformed the Baseline model, showing the worthiness of developing DL models for this multivariate forecasting task. Similar scores were achieved by the GRU and BiLSTM models with an average RMSE around 1241. The least performing DL method in our dataset was the LSTM model. Extending the architectures, hyperparameters range, lag values (i.e., test with less or more input time steps), dropout layers, for example, could probably improve our models and change the resulting best technique. However, we will focus our attention on a comparative analysis of differentially private DL methods in the next subsection and, thus, these possible extensions are left as future work.

\subsection{Differentially Private DL Forecasting Models} \label{sub:private_DL}

\noindent \textbf{Methods evaluated.} We consider two privacy-preserving ML settings presented in Subsection~\ref{sub:dp}, namely, input perturbation (IP) and gradient perturbation (GP). Thus, we selected only the DL method that performed the best with original data, i.e., BiGRU (cf. Table~\ref{tab:results_non_private}). We will use BiGRU[IP] and BiGRU[GP] to indicate a BiGRU trained under input and gradient perturbation, respectively. 

For the model selection stage, we first start with BiGRU[GP] since it allows defining a range of $\epsilon$, which is dependent on several hyperparameters of DP-SGD. For a fair comparison between both settings, we utilize the given range of $\epsilon$ to develop BiGRU[IP] models too. Notice, however, that in both scenarios, ($\epsilon,\delta$)-DP can be ensured to each time series data sample. On the other hand, this also means that the same user may have contributed to all $n_l=3120$ training samples and, thus, in the worst case, the sequential composition in Proposition~\ref{prop:sequential_composition} applies. With these elements in mind, we considered high privacy regimes ($\epsilon \ll 1$) such that the maximum $\check{ \epsilon} =  \sum_{i=1}^{n_l} \epsilon_i$ is compatible with real-world DP deployed systems~\cite[Table 2]{linkedin}. \textbf{This way, $\epsilon$ corresponds to the lower bound (the user appears in a single data point), and $\check{ \epsilon}$ represents the upper bound (the user appears in all data points).}

\noindent \textbf{BiGRU[GP] model selection.} In addition to standard hyperparameters $h_1$, $bs$, and $\eta$ (cf. Subsection~\ref{sub:non_private_DL}), we also included the TFP hyperparameters in the Bayesian optimization with $100$ iterations to minimize $loss=(RMSE_{avg} + RMSE_{std}) \times e^{\epsilon}$; the multiplicative factor $e^{\epsilon}$ is a penalization on high values of $\epsilon$, which varies depending on the hyperparameters used per iteration. More specifically, given the number of training samples $n_l=3120$, we fix the following hyperparameters: the number of epochs equal $100$, $num\_microbatches=5$, $noise\_multiplier$ equal $\{35, 70, 140, 500\}$, respectively, and $\delta=10^{-7}$, which respects $\sum_{i=1}^{n_l} \delta_i < 1/n_l$~\cite{dwork2014algorithmic}. This way, we varied $h_1$, $bs$, $\eta$, and $l2\_norm\_clip$ according to Table~\ref{tab:hyper_GP}, which exhibits the hyperparameters' search space, the final value used per BiGRU[GP] model, and the resulting privacy guarantee $\epsilon$ calculated with the \texttt{compute\_dp\_sgd\_privacy} function~\cite{tf_privacy}, and the overall $\check{ \epsilon} =  \sum_{i=1}^{n_l} \epsilon_i$. Lastly, all BiGRU[GP] models also used ``relu" as an activation function and were trained using the differentially private adam optimizer by minimizing the MAE loss function.

\begin{table*}[!ht]
    \centering
    \caption{Search space for standard and TFP hyperparameters, the best configuration per BiGRU[GP] model, the final privacy guarantee $\epsilon$ per time-series sample, and the maximum $\check{ \epsilon}$ following the sequential composition in Proposition~\ref{prop:sequential_composition}.}
    \begin{tabular}{c c c c c c}
    \hline
         \textbf{Hyperarameter's range} & \textbf{Step} & \textbf{BiGRU[GP]$_1$} & \textbf{BiGRU[GP]$_2$} & \textbf{BiGRU[GP]$_3$}  & \textbf{BiGRU[GP]$_4$}  \\\hline
         $h_1$: [25 -- 500]    & 25            &500       &425         &275      &475 \\
         $bs$: [5 -- 40]       & 5             &5         &5           &10        &5       \\
         $\eta$: [1e-5 -- 3e-3]  & --          &0.002229  &0.000455     &0.000291  &0.001235    \\
         $l2\_norm\_clip$ : \{1, 1.5, 2, 2.5\}  & --  &2.5      &2           &1      &2.5  \\
         $noise\_multiplier$ : \texttt{fixed}  & --     &35      &70    &140       &500    \\\hline  
         \multicolumn{2}{l}{\multirow{2}{*}{ \textbf{Privacy guarantee}}}  &  $\epsilon_1=0.0650$   &$\epsilon_2=0.0399$   &$\epsilon_3=0.0357$    &$\epsilon_4=0.0317$\\
         & &  $\check{ \epsilon}_1=202.8$     & $\check{ \epsilon}_2=124.488$       &  $\check{ \epsilon}_3=111.384$      & $\check{ \epsilon}_4=98.904$       \\
         \hline
    \end{tabular}
    \label{tab:hyper_GP}
\end{table*}

\noindent \textbf{BiGRU[IP] model selection.} We fix $\delta=10^{-7}$ and we apply the Gaussian mechanism~\cite{dwork2014algorithmic}, by varying $\epsilon$ according to Table~\ref{tab:hyper_GP}, \textbf{to the whole time series data, as it would be done if such system had been deployed in real life. The metrics, however, are computed in comparison with original raw time series data.} Because input perturbation allows using any post-processing techniques, we used the same model selection methodology as for non-private BiGRU models to select the best hyperparameters for BiGRU[IP] models. The resulting values per $\epsilon=[0.0650,0.0399,0.0357,0.0317]$, respectively, are: $\textrm{BiGRU[IP]}_1 :\{h_1=200, bs=5, \eta=0.001993\}$, $\textrm{BiGRU[IP]}_2 :\{h_1=275, bs=5, \eta=0.001182\}$,  $\textrm{BiGRU[IP]}_3 :\{h_1=200, bs=10, \eta=0.001333\}$, and $\textrm{BiGRU[IP]}_4 :\{h_1=200, bs=10, \eta=0.000842\}$.

\noindent \textbf{Privacy-preserving results and analysis.} Table~\ref{tab:results_private} presents the performance of differentially private BiGRU models trained under input and gradient perturbation regarding the RMSE and MAE metrics per region and the resulting mean values. We also included in Table~\ref{tab:results_private} the utility loss of differentially private BiGRU models in comparison with non-private ones, for both RMSE and MAE averaged metrics $\mathscr{E}$, calculated as:

\begin{equation}\label{eq:acc_loss}
    \mathscr{U} = \frac{ \mathscr{E}_{DP} - \mathscr{E}_{NP}} { \mathscr{E}_{NP}} \textrm{,}
\end{equation}

\noindent in which $\mathscr{E}_{NP}$ is the result of Non-Private BiGRU (cf. averaged metric values \textbf{in bold} from Table~\ref{tab:results_non_private}) and $\mathscr{E}_{DP}$ refers to the results of either BiGRU[GP] or BiGRU[IP] models. Indeed, Eq.~\eqref{eq:acc_loss} will be positive unless the differentially private model outperforms the non-private one (which is not the case in our results). 

\begin{table*}[!ht]
    \centering
    \caption{Performance of differentially private BiGRU models based on RMSE and MAE metrics per region and the resulting mean values. The last column $\mathscr{U}$ exhibits the utility loss of differentially private BiGRU models in comparison with non-private ones, for both RMSE and MAE averaged metrics expressed in $\%$.}
    \label{tab:results_private}
    \begin{tabular}{c c c c c c c c c c c} 
    \hline
    \textbf{$\epsilon,\check{ \epsilon}$ values}& \textbf{Model} & \textbf{Metric} & \textbf{R1} & \textbf{R2} & \textbf{R3}  & \textbf{R4}  & \textbf{R5}   & \textbf{R6}  & \textbf{Mean} & \textbf{$\mathscr{U}$} \\ 
    \hline
    \multirow{2}{*}{$\epsilon_1=0.0650$}&   \multirow{2}{*}{BiGRU[GP]$_1$}   & RMSE &2561.4    &1027.3    &1254.7  &866.7    &498.5    &1145.7    &1225.7  &0.9378 \\
                                      && MAE  &1973.4    &773.8    &925.     &644.1    &397.9    &781.5    &916.0  &0.9776   \\ \cline{2-11}
    \multirow{2}{*}{$\check{ \epsilon}_1=202.8$}&  \multirow{2}{*}{BiGRU[IP]$_1$} & RMSE &2600.9 &997.1  &1304.0  &852.7  &483.8 &1175.2  &1235.6  &1.7531  \\
                                &      & MAE  &1966.0    &737.5    &957.1    &645.4    &385.1    &821.1    &918.7  &1.2753  \\ \hline
    \multirow{2}{*}{$\epsilon_2=0.0399$}&   \multirow{2}{*}{BiGRU[GP]$_2$}   & RMSE &2600.2    &956.0     &1268.5    &841.5    &515.0  &1146.3  &\underline{1221.2} & \underline{0.5672} \\
                                      && MAE  &1978.9    &709.2    &944.4    &643.3    &417.4    &769.9    &910.5  &0.3713  \\ \cline{2-11}
    \multirow{2}{*}{$\check{ \epsilon}_2=124.488$}& \multirow{2}{*}{BiGRU[IP]$_2$}   & RMSE &2592.2  &978.4  &1251.5  &854.2   &495.6    &1158.6  &\underline{1221.8}  &\underline{0.6166}  \\
                                &      & MAE  &1986.1    &737.1    &910.9    &653.9    &393.2    &813.9    & 915.9  &0.9666  \\ \hline   
    \multirow{2}{*}{$\epsilon_3=0.0357$}&   \multirow{2}{*}{BiGRU[GP]$_3$}   & RMSE &2580.5  &990.0    &1268.5    &854.5    &504.9    &1154.3  &\textbf{1225.5}  &\textbf{0.9213}  \\
                                      && MAE  &1938.8    &753.0    &942.8    &659.7    &406.3    &773.6    &912.4  &0.5808  \\ \cline{2-11}
    \multirow{2}{*}{$\check{ \epsilon}_3=111.384$}& \multirow{2}{*}{BiGRU[IP]$_3$}   & RMSE &2587.8 &1004.7 &1262.3  &843.2    &512.8    &1186.2  &\textbf{1232.9}  &\textbf{1.5307}  \\
                                &      & MAE  &1963.1    &755.8    &957.5    &636.9    &414.6    &811.8    & 923.3  &1.7824  \\ \hline   
    \multirow{2}{*}{$\epsilon_4=0.0317$}&   \multirow{2}{*}{BiGRU[GP]$_4$}   & RMSE &2560.8    &978.3    &1322.5    &836.1   &494.4    &1195.4 &1231.3  &1.3990  \\
                                      && MAE  &1956.2    &715.1    &989.2    &633.6   &392.0    &821.6   &917.9   &1.1871  \\ \cline{2-11}
    \multirow{2}{*}{$\check{ \epsilon}_4=98.904$}& \multirow{2}{*}{BiGRU[IP]$_4$} & RMSE &2562.2    &1012.2    &1351.2    &862.9  &533.5  &1168.8  &1248.4  &2.8072 \\
                                &      & MAE  &1955.6    &756.8    &1027.6  &650.1    &423.9    &826.8      &940.2  &3.6454 \\ \hline   
    \end{tabular}
\end{table*}

We remarked in our experiments that since there is a sufficient number of users per time series sample (cf. Table~\ref{tab:statistics_data}), it was still possible to make accurate forecasts in both privacy-preserving ML settings with the experimented range of ($\epsilon,\delta$)-DP. Indeed, one can notice that all differentially private BiGRU models achieved averaged RMSE lower than 1250, in which the worst result achieved by BiGRU[IP]$_4$ is just $2.8\%$ less precise than the non-private BiGRU model. What is more, in both gradient and input perturbation settings, differentially private BiGRU models achieved smaller error metrics than non-private LSTM, BiLSTM, and GRU models (cf. Table~\ref{tab:results_non_private}). For instance, both BiGRU[GP]$_2$ and BiGRU[IP]$_2$ reached similar scores in comparison with the non-private BiGRU model, with a loss of performance of about $0.57\%$ and $0.62\%$, respectively. These results are highlighted in \underline{underlined} font, which represents our best results in terms of utility, with differentially private BiGRU models. 

Interestingly, the accuracy (measured with the RMSE metric) of differentially private BiGRU models did not necessarily decrease according to more strict $\epsilon$, i.e., lower values. One can note that results with $\epsilon_2$ and $\epsilon_3$ were more accurate than with $\epsilon_1$. This way, in terms of a good \textit{privacy-utility trade-off}, both BiGRU[GP]$_3$ ($0.92\%$ less accurate) and BiGRU[IP]$_3$ ($1.53\%$ less accurate) presented good metrics scores while satisfying a low value of $\epsilon$, with results highlighted \textbf{in bold}. Indeed, in the worst-case scenario, a user that was present in each data point would have leaked $\check{ \epsilon}_3=111.384$ at the end of 65 days (i.e., $\epsilon \sim 1.7$ per day), which follows real-world DP systems deployed by industry nowadays~\cite[Section 8.4]{linkedin}.

\begin{figure*}[!ht]
    \centering
    \includegraphics[width=0.775\linewidth]{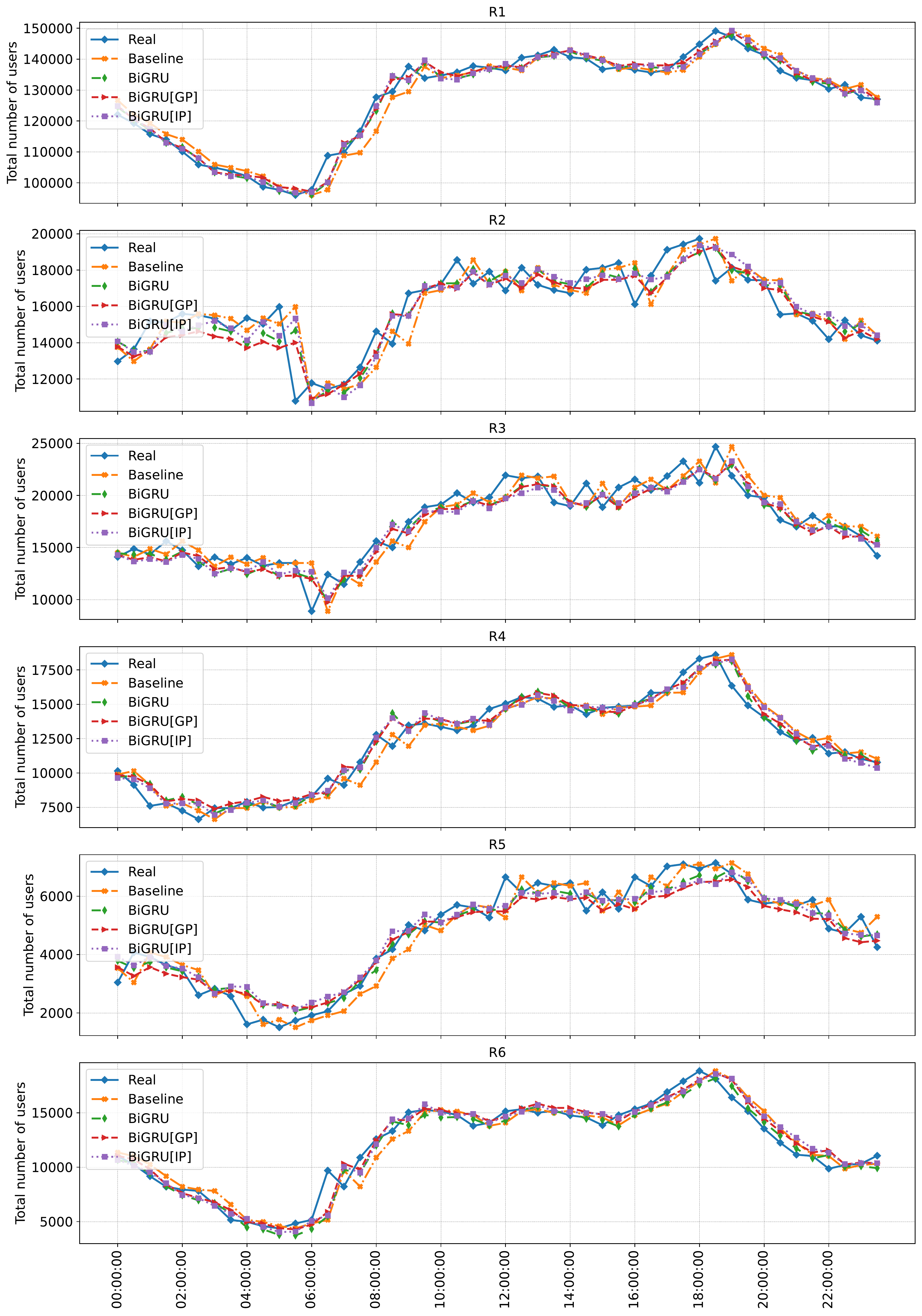}
    \caption{Multivariate time series forecast for the last day of the test set for the number of users per coarse region (R1 -- R6) by the following models: Baseline, non-private BiGRU, BiGRU[GP]$_3$, and BiGRU[IP]$_3$.}
    \label{fig:results_pred}
\end{figure*}

Lastly, Fig.~\ref{fig:results_pred} illustrates for each region forecasting results for the last day of our testing set, which includes the real number of people and the predicted ones by the following models: Baseline, non-private BiGRU, BiGRU[GP]$_3$, and BiGRU[IP]$_3$. As one can notice, similar forecasting results were achieved by both non-private and DP-based BiGRU models, which clearly outperforms the Baseline model. 

\section{Discussion and Related Work} \label{sec:discussion}

Mobile phone CDRs have been largely used to analyze human mobility in several contexts, e.g., the spread of infectious diseases~\cite{ebola,deAlarcon2021}, natural disasters~\cite{Hong2018,Dujardin2020}, tourism~\cite{fluxvision1}, and so on. However, on analyzing mobility data, de Montjoye et al.~\cite{deMontjoye2013} show that humans follow particular patterns, which allows predicting human mobility with high accuracy. For instance, in a CDRs dataset of $1.5$ million users, the authors showed that $95\%$ of this population can be re-identified using four approximate locations and their timestamps. Indeed, the uniqueness of mobility data has also been studied in~\cite{Murakami2021}, for example, in which the authors concluded that location trace has higher identifiability than a face matcher in a partial-knowledge model. 

For this reason, MNOs tend to publish aggregated mobility data~\cite{deAlarcon2021,Xu2017,Vespe2021,Tu2018,fluxvision1}, which provides some form of anonymity-based protection. However, as recent studies have shown, even aggregated mobility data (e.g., heatmaps) can be subject to membership inference attacks~\cite{Pyrgelis2017,Pyrgelis2020} and users' trajectory recovery attack~\cite{Tu2018,Xu2017}. More precisely, the authors in~\cite{Tu2018,Xu2017} showed that their attack reaches accuracies as high as $73\% \sim 91\%$. Therefore, it is vital to design privacy-preserving techniques that allow analyzing human mobility~\cite{deMontjoye2018}. 

Moreover, along with collecting time-series data, extracting meaningful forecasts is also of great interest. Time series forecasting has been a key area of ML research and application across many domains, e.g., medicine~\cite{Rahimi2021}, finance~\cite{Sezer2020}, electrical power~\cite{Stefenon2020,RodriguesMoreno2020}, mobility~\cite{luca2020deep}, and so on. However, even ML models trained with raw data can also indirectly reveal sensitive information~\cite{Carlini2021,Song2017,Carlini2019,shokri2017membership}, in particular, RNNs~\cite{yang2021privacy}. To protect ML models against such threats, under the state-of-the-art DP guarantee~\cite{Dwork2006,dwork2014algorithmic}, there exist some privacy-preserving ML alternatives adopted in the literature, e.g., input~\cite{Chen2020,Chamikara2020,Eibl2018,Imtiaz2020,Arcolezi2021c,Arcolezi2021b}, gradient~\cite{DL_DP,tf_privacy,UstundagSoykan2019,pytorch_privacy,Shokri2015}, and objective perturbation~\cite{chaudhuri2011differentially}. 

The contribution of our research is significant for those involved in urban planning and decision-making~\cite{deMontjoye2018}, providing a solution to the human mobility multivariate forecast problem through RNNs and differentially private BiGRUs. In addition, we point out the research community to the Github page mentioned in the introduction section, in which we release the mobility dataset used in this paper for further experimentation with time series, machine learning, and privacy-preserving methods. The related literature to our work includes the generation of synthetic mobility data~\cite{Ouyang2018,Mir2013,Arcolezi2020}, the development of Markov models to infer travelers’ activity pattern~\cite{Yin2018}, and the development of privacy-preserving methods to analyze CDRs-based data~\cite{app_blip,Arcolezi2021,Acs2014}. Besides, the work in~\cite{luca2020deep} surveys non-private deep learning applications to mobility datasets in general. Concerning differentially private deep learning, one can find the application of gradient perturbation-based DL models for load forecasting~\cite{UstundagSoykan2019}, an evaluation of differentially private DL models in federated learning for health stream forecasting~\cite{Imtiaz2020}, the proposal of locally differentially private DL architectures~\cite{Chamikara2020}, practical libraries for differentially private DL~\cite{tf_privacy,pytorch_privacy}, and theoretical research works~\cite{DL_DP,Shokri2015,Chen2020}.

In this work, accurate multivariate forecasts were achieved with four non-private RNNs (i.e., LSTM, GRU, BiLSTM, and BiGRU), with BiGRU standing out among the four methods. Thus, this paper further evaluated both input and gradient perturbation settings to forecast multivariate aggregated mobility time series data using the BiGRU neural network. Between both input and gradient perturbation settings, although not measured, BiGRU[GP] models took more time to execute than BiGRU[IP] models due to DP-SGD. In terms of accuracy, BiGRU[GP] models consistently outperformed BiGRU[IP] models for the same ($\epsilon, \delta$)-DP privacy level in our experiments. One reason for such result is because the \textit{input data perturbation} setting adds DP guarantees to each time series point in the data, trading privacy with utility. This is indeed one fundamental problem in DP theory~\cite{dwork2014algorithmic} in which local DP algorithms has low utility in comparison with centralized DP algorithms. On the other hand, as BiGRU[GP] is trained over non-DP time-series data, the mobility dataset is still subject to data leakage~\cite{data_breaches}, and consequently, membership inference attacks~\cite{Pyrgelis2017,Pyrgelis2020}, and users' trajectory recovery attacks~\cite{Tu2018,Xu2017}, which requires strong security measures. Therefore, training ML models over differentially private multivariate time series mobility data provides the best privacy-utility trade-off. In practice, the input-perturbation setting allows applying centralized DP mechanisms in the final aggregate data (e.g., Laplace mechanism~\cite{Dwork2006}, Gaussian mechanism~\cite{dwork2014algorithmic}), essentially refreshing the privacy budget $\epsilon$ on a regular basis, and using the published data for any purpose (cf. Proposition~\ref{prop:post_processing}).

Finally, some limitations and prospective directions of this paper are described in the following. For differentially private BiGRU models, we only provided lower $\epsilon$ and upper $\check{\epsilon}$ bounds for the privacy guarantee of each sample in the time-series data. Using, however, advanced composition theorems~\cite{dwork2014algorithmic} to account for the final privacy budget for each user was out of the scope of this paper. Indeed, CDRs are event-based~\cite{Oliver2020,Dujardin2020}, which means that data are only available when users actively make phone calls (or connect to the internet, or send SMS). This way, there may have users who make several calls (e.g., business people) and, thus, have higher values of $\check{\epsilon}$, while some groups do not, e.g., poor people. Besides, although the developed DL models outperform the Baseline model ($\textbf{x}_{t+1}=\textbf{x}_t$), there is plenty of room for improvements to be carried out on hyperparameters optimization, data scaling, the number of lag values, etc. For instance, some high-peak values were missed by both non-private and DP-based DL models (see Figs.~\ref{fig:results_pred_non_private} and~\ref{fig:results_pred}). In addition, we fixed the number of lagged values to 6 to predict a single step-ahead in the future (i.e., the forecasting horizon), in which the former can be tuned for performance improvement and the latter can be increased for multi-step forecasting tasks. Thus, besides the aforementioned directions, for future work, we suggest and intend to investigate a more complex DL architecture to improve the results of DL models proposed in this paper for this multivariate time series forecasting task. Lastly, investigating the data leakage through membership inference attacks~\cite{shokri2017membership,yang2021privacy} of both privacy-preserving ML settings is also a prospective and intended direction. 

\section{Conclusion} \label{sec:conclusion}

This paper provides the first comparative evaluation of differentially private DL models in both input and gradient perturbation settings to forecast multivariate aggregated mobility time series data. Experiments were first carried out with four non-private DL models (i.e., LSTM, GRU, BiLSTM, and BiGRU). The BiGRU model best fitted our data and, thus, it was selected for building differentially private DL models. Under gradient and input perturbation settings, i.e., BiGRU[GP] and BiGRU[IP], respectively, four values of $\epsilon \ll 1$ were evaluated. As shown in the results, differentially private BiGRU models achieve nearly the same performance as non-private BiGRU models, with loss in performance varying between $0.57\%$ to $2.8\%$ (for the RMSE metric). Thus, we conclude that it is still possible to have accurate multivariate forecasts in both privacy-preserving ML settings. More specifically, although the gradient perturbation setting preserved more accuracy than the input perturbation setting, input perturbation guarantees stronger privacy protection (i.e., both for the ML model and for the data itself), thus providing the best privacy-utility trade-off.

\begin{acknowledgements}
\noindent This work was supported by the EIPHI-BFC Graduate School (contract ``ANR-17-EURE-0002") and by the Region of Bourgogne Franche-Comt\'e CADRAN Project. The work of H\'eber H. Arcolezi has been partially supported by the ERC project Hypatia, grant agreement Nº 835294. The authors would also like to thank the \enquote{Orange Application for Business} team for their continuous collaborations and useful feedback. All computations have been performed on the \enquote{M\'esocentre de Calcul de Franche-Comt\'e}.
\end{acknowledgements}

\section*{Declarations}

\subsection*{Funding}
No funding was received to assist with the preparation of this manuscript.

\subsection*{Conflict of Interest}
The authors have no conflicts of interest to declare that are relevant to the content of this article.

\subsection*{Availability of data and material}
Data can be accessed on the following Github page (\url{https://github.com/hharcolezi/ldp-protocols-mobility-cdrs}).

\subsection*{Code availability}
Codes are available on the following Github page (\url{https://github.com/hharcolezi/ldp-protocols-mobility-cdrs}).

\subsection*{Research involving Human Participants and/or Animals}
This article does not contain any studies with human or animal subjects performed by any of the authors.

\subsection*{Informed Consent} 
As this article does not contain any studies with human participants or animals, the informed consent is not applicable.

\bibliographystyle{spmpsci}      
\bibliography{ms.bib}   

\begin{thebibliography}{10}
\providecommand{\url}[1]{{#1}}
\providecommand{\urlprefix}{URL }
\expandafter\ifx\csname urlstyle\endcsname\relax
  \providecommand{\doi}[1]{DOI~\discretionary{}{}{}#1}\else
  \providecommand{\doi}{DOI~\discretionary{}{}{}\begingroup
  \urlstyle{rm}\Url}\fi

\bibitem{lockdown}
Confinements liés à la pandémie de {COVID-19} en france.
\newblock Available online:
  \url{https://fr.wikipedia.org/wiki/Confinements_li\%C3\%A9s\_\%C3\%A0\_la_pand\%C3\%A9mie\_de\_Covid-19\_en\_France}
  (accessed on 11 July 2021)

\bibitem{CNIL}
Commission nationale de l'informatique et des libertés ({CNIL}) (1978).
\newblock Available online: \url{https://www.cnil.fr/en/home} (accessed on 04
  July 2021)

\bibitem{GDPR}
General data protection regulation ({GDPR}) (2018).
\newblock Available online: \url{https://gdpr-info.eu/} (accessed on 04 July
  2021)

\bibitem{DL_DP}
Abadi, M., Chu, A., Goodfellow, I., McMahan, H.B., Mironov, I., Talwar, K.,
  Zhang, L.: Deep learning with differential privacy.
\newblock CCS '16, p. 308–318. Association for Computing Machinery, New York,
  NY, USA (2016).
\newblock \doi{10.1145/2976749.2978318}

\bibitem{Acs2014}
Acs, G., Castelluccia, C.: A case study: Privacy preserving release of
  spatio-temporal density in {Paris}.
\newblock In: Proceedings of the 20th {ACM} {SIGKDD} international conference
  on Knowledge discovery and data mining - {KDD} {\textquotesingle}14. {ACM}
  Press (2014).
\newblock \doi{10.1145/2623330.2623361}

\bibitem{aktay2020google}
Aktay, A., Bavadekar, S., Cossoul, G., Davis, J., Desfontaines, D., Fabrikant,
  A., Gabrilovich, E., Gadepalli, K., Gipson, B., Guevara, M., et~al.: Google
  {COVID-19} community mobility reports: anonymization process description
  (version 1.1).
\newblock arXiv preprint arXiv:2004.04145  (2020)

\bibitem{app_blip}
Alaggan, M., Gambs, S., Matwin, S., Tuhin, M.: Sanitization of call detail
  records via differentially-private bloom filters.
\newblock In: Data and Applications Security and Privacy {XXIX}, pp. 223--230.
  Springer International Publishing (2015)

\bibitem{deAlarcon2021}
de~Alarcon, P.A., Salevsky, A., Gheti-Kao, D., Rosalen, W., Duarte, M.C.,
  Cuervo, C., Mu{\~{n}}oz, J.J., Pascual, J.M., Schurig, M., Tre{\ss}, T.,
  Diaz, E., de~la Cuesta, C., Frias-Martinez, E.: The contribution of telco
  data to fight the {COVID}-19 pandemic: Experience of telefonica throughout
  its footprint.
\newblock Data {\&} Policy \textbf{3}, e7 (2021).
\newblock \doi{10.1017/dap.2021.6}

\bibitem{Arcolezi2021b}
Arcolezi, H.H., Cerna, S., Couchot, J.F., Guyeux, C., Makhoul, A.:
  Privacy-preserving prediction of victim’s mortality and their need for
  transportation to health facilities.
\newblock IEEE Transactions on Industrial Informatics \textbf{18}(8),
  5592--5599 (2022).
\newblock \doi{10.1109/TII.2021.3123588}

\bibitem{Arcolezi2021c}
Arcolezi, H.H., Cerna, S., Guyeux, C., Couchot, J.F.: Preserving
  geo-indistinguishability of the emergency scene to predict ambulance response
  time.
\newblock Mathematical and Computational Applications \textbf{26}(3) (2021).
\newblock \doi{10.3390/mca26030056}

\bibitem{Arcolezi2020}
Arcolezi, H.H., Couchot, J.F., Baala, O., Contet, J.M., Al~Bouna, B., Xiao, X.:
  Mobility modeling through mobile data: generating an optimized and open
  dataset respecting privacy.
\newblock In: 2020 International Wireless Communications and Mobile Computing
  (IWCMC), pp. 1689--1694 (2020).
\newblock \doi{10.1109/IWCMC48107.2020.9148138}

\bibitem{Arcolezi2021}
Arcolezi, H.H., Couchot, J.F., Bouna, B.A., Xiao, X.: Longitudinal collection
  and analysis of mobile phone data with local differential privacy.
\newblock In: M.~Friedewald, S.~Schiffner, S.~Krenn (eds.) Privacy and Identity
  Management, pp. 40--57. Springer International Publishing, Cham (2021).
\newblock \doi{10.1007/978-3-030-72465-8_3}

\bibitem{hyperopt2013}
Bergstra, J., Yamins, D., Cox, D.D.: Making a science of model search:
  Hyperparameter optimization in hundreds of dimensions for vision
  architectures.
\newblock In: Proceedings of the 30th International Conference on International
  Conference on Machine Learning, ICML'13, p. I–115–I–123. JMLR (2013)

\bibitem{Blondel2015}
Blondel, V.D., Decuyper, A., Krings, G.: A survey of results on mobile phone
  datasets analysis.
\newblock {EPJ} Data Science \textbf{4}(1), 10 (2015).
\newblock \doi{10.1140/epjds/s13688-015-0046-0}

\bibitem{Buckee2020}
Buckee, C.O., Balsari, S., Chan, J., Crosas, M., Dominici, F., Gasser, U.,
  Grad, Y.H., Grenfell, B., Halloran, M.E., Kraemer, M.U.G., Lipsitch, M.,
  Metcalf, C.J.E., Meyers, L.A., Perkins, T.A., Santillana, M., Scarpino, S.V.,
  Viboud, C., Wesolowski, A., Schroeder, A.: Aggregated mobility data could
  help fight {COVID}-19.
\newblock Science \textbf{368}(6487), 145--146 (2020).
\newblock \doi{10.1126/science.abb8021}

\bibitem{Carlini2019}
Carlini, N., Liu, C., Erlingsson, {\'U}., Kos, J., Song, D.: The secret sharer:
  Evaluating and testing unintended memorization in neural networks.
\newblock In: 28th USENIX Security Symposium (USENIX Security 19), pp.
  267--284. USENIX Association, Santa Clara, CA (2019)

\bibitem{Carlini2021}
Carlini, N., Tram{\`e}r, F., Wallace, E., Jagielski, M., Herbert-Voss, A., Lee,
  K., Roberts, A., Brown, T., Song, D., Erlingsson, {\'U}., Oprea, A., Raffel,
  C.: Extracting training data from large language models.
\newblock In: 30th USENIX Security Symposium (USENIX Security 21), pp.
  2633--2650. USENIX Association (2021)

\bibitem{chaudhuri2011differentially}
Chaudhuri, K., Monteleoni, C., Sarwate, A.D.: Differentially private empirical
  risk minimization.
\newblock Journal of Machine Learning Research \textbf{12}(29), 1069--1109
  (2011)

\bibitem{Chen2020}
Chen, S., Fu, A., Shen, J., Yu, S., Wang, H., Sun, H.: {RNN}-{DP}: A new
  differential privacy scheme base on recurrent neural network for dynamic
  trajectory privacy protection.
\newblock Journal of Network and Computer Applications \textbf{168}, 102736
  (2020).
\newblock \doi{10.1016/j.jnca.2020.102736}

\bibitem{GRU}
Chung, J., Gulcehre, C., Cho, K., Bengio, Y.: Empirical evaluation of gated
  recurrent neural networks on sequence modeling.
\newblock In: NIPS 2014 Workshop on Deep Learning (2014)

\bibitem{Dujardin2020}
Dujardin, S., Jacques, D., Steele, J., Linard, C.: Mobile phone data for urban
  climate change adaptation: Reviewing applications, opportunities and key
  challenges.
\newblock Sustainability \textbf{12}(4), 1501 (2020).
\newblock \doi{10.3390/su12041501}

\bibitem{Dwork2006}
Dwork, C., McSherry, F., Nissim, K., Smith, A.: Calibrating noise to
  sensitivity in private data analysis.
\newblock In: Theory of Cryptography, pp. 265--284. Springer Berlin Heidelberg
  (2006).
\newblock \doi{10.1007/11681878_14}

\bibitem{dwork2014algorithmic}
Dwork, C., Roth, A.: The algorithmic foundations of differential privacy.
\newblock Foundations and Trends{\textregistered} in Theoretical Computer
  Science \textbf{9}(3--4), 211--407 (2014)

\bibitem{Eibl2018}
Eibl, G., Bao, K., Grassal, P.W., Bernau, D., Schmeck, H.: The influence of
  differential privacy on short term electric load forecasting.
\newblock Energy Informatics \textbf{1}(S1) (2018).
\newblock \doi{10.1186/s42162-018-0025-3}

\bibitem{europecom2020}
European-Commission: Commission recommendation (eu) 2020/518 of 8 april 2020 on
  a common union toolbox for the use of technology and data to combat and exit
  from the {COVID-19} crisis, in particular concerning mobile applications and
  the use of anonymised mobility data.
\newblock Available online:
  \url{https://eur-lex.europa.eu/eli/reco/2020/518/oj} (accessed on 04 July
  2021)

\bibitem{Hewamalage2021}
Hewamalage, H., Bergmeir, C., Bandara, K.: Recurrent neural networks for time
  series forecasting: Current status and future directions.
\newblock International Journal of Forecasting \textbf{37}(1), 388--427 (2021).
\newblock \doi{10.1016/j.ijforecast.2020.06.008}

\bibitem{LSTM}
Hochreiter, S., Schmidhuber, J.: Long short-term memory.
\newblock Neural computation \textbf{9}(8), 1735--1780 (1997)

\bibitem{Hong2018}
Hong, L., Lee, M., Mashhadi, A., Frias-Martinez, V.: Towards understanding
  communication behavior changes during floods using cell phone data.
\newblock In: Lecture Notes in Computer Science, pp. 97--107. Springer
  International Publishing (2018).
\newblock \doi{10.1007/978-3-030-01159-8_9}

\bibitem{Imtiaz2020}
Imtiaz, S., Horchidan, S.F., Abbas, Z., Arsalan, M., Chaudhry, H.N., Vlassov,
  V.: Privacy preserving time-series forecasting of user health data streams.
\newblock In: 2020 {IEEE} International Conference on Big Data (Big Data).
  {IEEE} (2020).
\newblock \doi{10.1109/bigdata50022.2020.9378186}

\bibitem{luca2020deep}
Luca, M., Barlacchi, G., Lepri, B., Pappalardo, L.: A survey on deep learning
  for human mobility.
\newblock ACM Comput. Surv. \textbf{55}(1) (2021).
\newblock \doi{10.1145/3485125}

\bibitem{Chamikara2020}
Mahawaga~Arachchige, P.C., Bertok, P., Khalil, I., Liu, D., Camtepe, S.,
  Atiquzzaman, M.: Local differential privacy for deep learning.
\newblock IEEE Internet of Things Journal \textbf{7}(7), 5827--5842 (2020).
\newblock \doi{10.1109/JIOT.2019.2952146}

\bibitem{data_breaches}
McCandless, D., Evans, T., Quick, M., Hollowood, E., Miles, C., Hampson, D.,
  Geere, D.: World's biggest data breaches \& hacks.
\newblock
  \url{https://www.informationisbeautiful.net/visualizations/worlds-biggest-data-breaches-hacks/}
  (2021).
\newblock Online; accessed 11 March 2021

\bibitem{tf_privacy}
McMahan, H.B., Andrew, G., Erlingsson, U., Chien, S., Mironov, I., Papernot,
  N., Kairouz, P.: A general approach to adding differential privacy to
  iterative training procedures.
\newblock In: Advances in Neural Information Processing Systems (NeurIPS)
  Workshop on Privacy Preserving Machine Learning (2018)

\bibitem{Mir2013}
Mir, D.J., Isaacman, S., Caceres, R., Martonosi, M., Wright, R.N.:
  {DP}-{WHERE}: Differentially private modeling of human mobility.
\newblock In: 2013 {IEEE} International Conference on Big Data. {IEEE} (2013).
\newblock \doi{10.1109/bigdata.2013.6691626}

\bibitem{deMontjoye2018}
de~Montjoye, Y.A., Gambs, S., Blondel, V., Canright, G., de~Cordes, N.,
  Deletaille, S., Eng{\o}-Monsen, K., Garcia-Herranz, M., Kendall, J., Kerry,
  C., Krings, G., Letouz{\'{e}}, E., Luengo-Oroz, M., Oliver, N., Rocher, L.,
  Rutherford, A., Smoreda, Z., Steele, J., Wetter, E., Pentland,
  A.{\textquotedblleft}., Bengtsson, L.: On the privacy-conscientious use of
  mobile phone data.
\newblock Scientific Data \textbf{5}(1), 180286 (2018).
\newblock \doi{10.1038/sdata.2018.286}

\bibitem{deMontjoye2013}
de~Montjoye, Y.A., Hidalgo, C.A., Verleysen, M., Blondel, V.D.: Unique in the
  crowd: The privacy bounds of human mobility.
\newblock Scientific Reports \textbf{3}(1), 1376 (2013).
\newblock \doi{10.1038/srep01376}

\bibitem{RodriguesMoreno2020}
Moreno, S.R., da~Silva, R.G., Mariani, V.C., dos Santos~Coelho, L.: Multi-step
  wind speed forecasting based on hybrid multi-stage decomposition model and
  long short-term memory neural network.
\newblock Energy Conversion and Management \textbf{213}, 112869 (2020).
\newblock \doi{10.1016/j.enconman.2020.112869}

\bibitem{Murakami2021}
Murakami, T., Takahashi, K.: Toward evaluating re-identification risks in the
  local privacy model.
\newblock Transactions on Data Privacy \textbf{14}, 79--116 (2021)

\bibitem{Oliver2020}
Oliver, N., Lepri, B., Sterly, H., Lambiotte, R., Deletaille, S., Nadai, M.D.,
  Letouzé, E., Salah, A.A., Benjamins, R., Cattuto, C., Colizza, V.,
  de~Cordes, N., Fraiberger, S.P., Koebe, T., Lehmann, S., Murillo, J.,
  Pentland, A., Pham, P.N., Pivetta, F., Saramäki, J., Scarpino, S.V.,
  Tizzoni, M., Verhulst, S., Vinck, P.: Mobile phone data for informing public
  health actions across the {COVID-19} pandemic life cycle.
\newblock Science Advances \textbf{6}(23), eabc0764 (2020).
\newblock \doi{10.1126/sciadv.abc0764}

\bibitem{fluxvision1}
Orange-Business-Services: Flux vision: real time statistics on mobility
  patterns (2013).
\newblock Available online:
  \url{https://www.orange-business.com/en/products/flux-vision} (accessed on 01
  July 2021)

\bibitem{Ouyang2018}
Ouyang, K., Shokri, R., Rosenblum, D.S., Yang, W.: A non-parametric generative
  model for human trajectories.
\newblock IJCAI'18, p. 3812–3817. AAAI Press (2018)

\bibitem{Pyrgelis2017}
Pyrgelis, A., Troncoso, C., Cristofaro, E.D.: What does the crowd say about
  you? evaluating aggregation-based location privacy.
\newblock Proceedings on Privacy Enhancing Technologies \textbf{2017}(4),
  156--176 (2017).
\newblock \doi{10.1515/popets-2017-0043}

\bibitem{Pyrgelis2020}
Pyrgelis, A., Troncoso, C., Cristofaro, E.D.: Measuring membership privacy on
  aggregate location time-series.
\newblock In: Abstracts of the 2020 {SIGMETRICS}/Performance Joint
  International Conference on Measurement and Modeling of Computer Systems, pp.
  1--28. {ACM} (2020).
\newblock \doi{10.1145/3393691.3394200}

\bibitem{Rahimi2021}
Rahimi, I., Chen, F., Gandomi, A.H.: A review on {COVID}-19 forecasting models.
\newblock Neural Computing and Applications  (2021).
\newblock \doi{10.1007/s00521-020-05626-8}

\bibitem{linkedin}
Rogers, R., Subramaniam, S., Peng, S., Durfee, D., Lee, S., Kancha, S.K.,
  Sahay, S., Ahammad, P.: Linkedin’s audience engagements {API}: A privacy
  preserving data analytics system at scale.
\newblock Journal of Privacy and Confidentiality \textbf{11}(3) (2021).
\newblock \doi{10.29012/jpc.782}

\bibitem{BI_RNN}
Schuster, M., Paliwal, K.: Bidirectional recurrent neural networks.
\newblock {IEEE} Transactions on Signal Processing \textbf{45}(11), 2673--2681
  (1997).
\newblock \doi{10.1109/78.650093}

\bibitem{Sezer2020}
Sezer, O.B., Gudelek, M.U., Ozbayoglu, A.M.: Financial time series forecasting
  with deep learning : A systematic literature review: 2005{\textendash}2019.
\newblock Applied Soft Computing \textbf{90}, 106181 (2020).
\newblock \doi{10.1016/j.asoc.2020.106181}

\bibitem{Shokri2015}
Shokri, R., Shmatikov, V.: Privacy-preserving deep learning.
\newblock CCS '15, p. 1310–1321. Association for Computing Machinery, New
  York, NY, USA (2015).
\newblock \doi{10.1145/2810103.2813687}

\bibitem{shokri2017membership}
Shokri, R., Stronati, M., Song, C., Shmatikov, V.: Membership inference attacks
  against machine learning models.
\newblock In: 2017 IEEE symposium on security and privacy (SP), pp. 3--18. IEEE
  (2017).
\newblock \doi{10.1109/sp.2017.41}

\bibitem{Song2017}
Song, C., Ristenpart, T., Shmatikov, V.: Machine learning models that remember
  too much.
\newblock CCS '17, p. 587–601. Association for Computing Machinery, New York,
  NY, USA (2017).
\newblock \doi{10.1145/3133956.3134077}

\bibitem{UstundagSoykan2019}
Soykan, E.U., Bilgin, Z., Ersoy, M.A., Tomur, E.: Differentially private deep
  learning for load forecasting on smart grid.
\newblock In: 2019 {IEEE} Globecom Workshops ({GC} Wkshps), pp. 1--6. {IEEE}
  (2019).
\newblock \doi{10.1109/gcwkshps45667.2019.9024520}

\bibitem{Stefenon2020}
Stefenon, S.F., Ribeiro, M.H.D.M., Nied, A., Mariani, V.C., dos Santos~Coelho,
  L., da~Rocha, D.F.M., Grebogi, R.B., de~Barros~Ruano, A.E.: Wavelet group
  method of data handling for fault prediction in electrical power insulators.
\newblock International Journal of Electrical Power {\&} Energy Systems
  \textbf{123}, 106269 (2020).
\newblock \doi{10.1016/j.ijepes.2020.106269}

\bibitem{Tu2018}
Tu, Z., Xu, F., Li, Y., Zhang, P., Jin, D.: A new privacy breach: User
  trajectory recovery from aggregated mobility data.
\newblock {IEEE}/{ACM} Transactions on Networking \textbf{26}(3), 1446--1459
  (2018).
\newblock \doi{10.1109/tnet.2018.2829173}

\bibitem{Vespe2021}
Vespe, M., Iacus, S.M., Santamaria, C., Sermi, F., Spyratos, S.: On the use of
  data from multiple mobile network operators in europe to fight {COVID}-19.
\newblock Data {\&} Policy \textbf{3}, e8 (2021).
\newblock \doi{10.1017/dap.2021.9}

\bibitem{ebola}
Wesolowski, A., Buckee, C.O., Bengtsson, L., Wetter, E., Lu, X., Tatem, A.J.:
  Commentary: Containing the ebola outbreak - the potential and challenge of
  mobile network data.
\newblock {PLoS} Currents  (2014).
\newblock \doi{10.1371/currents.outbreaks.0177e7fcf52217b8b634376e2f3efc5e}

\bibitem{who_annouces_pandemic}
World-Health-Organization: {WHO} announces {COVID-19} outbreak a pandemic.
\newblock Available online:
  \url{https://www.euro.who.int/en/health-topics/health-emergencies/coronavirus-covid-19/news/news/2020/3/who-announces-covid-19-outbreak-a-pandemic}
  (accessed on 07 September 2020)

\bibitem{Xu2017}
Xu, F., Tu, Z., Li, Y., Zhang, P., Fu, X., Jin, D.: Trajectory recovery from
  {ASH}.
\newblock In: Proceedings of the 26th International Conference on World Wide
  Web, pp. 1241--1250. International World Wide Web Conferences Steering
  Committee (2017).
\newblock \doi{10.1145/3038912.3052620}

\bibitem{yang2021privacy}
Yang, Y., Gohari, P., Topcu, U.: On the privacy risks of deploying recurrent
  neural networks in machine learning.
\newblock arXiv preprint arXiv:2110.03054  (2021)

\bibitem{Yin2018}
Yin, M., Sheehan, M., Feygin, S., Paiement, J.F., Pozdnoukhov, A.: A generative
  model of urban activities from cellular data.
\newblock IEEE Transactions on Intelligent Transportation Systems
  \textbf{19}(6), 1682--1696 (2018).
\newblock \doi{10.1109/TITS.2017.2695438}

\bibitem{pytorch_privacy}
Yousefpour, A., Shilov, I., Sablayrolles, A., Testuggine, D., Prasad, K.,
  Malek, M., Nguyen, J., Ghosh, S., Bharadwaj, A., Zhao, J., Cormode, G.,
  Mironov, I.: Opacus: User-friendly differential privacy library in pytorch.
\newblock In: NeurIPS 2021 Workshop Privacy in Machine Learning (2021)

\end{thebibliography}

\end{document}